\documentclass{article}
\usepackage{spconf,amsmath,graphicx}
\usepackage{multirow}

\def\vsp{{\vspace{-0.5em}}}
\usepackage{microtype}

\title{Lexico-acoustic Neural-based Models for Dialog Act Classification}
\name{Daniel Ortega \qquad Ngoc Thang Vu}
\address{Institute for Natural Language Processing (IMS)\\
	University of Stuttgart, Germany\\
  {\tt \{daniel.ortega, thang.vu\}@ims.uni-stuttgart.de}}
\begin{document}
%
\maketitle
\begin{abstract}
Recent works have proposed neural models for dialog act classification in spoken dialogs. 
However, they have not explored the role and the usefulness of acoustic information. 
We propose a neural model that processes both lexical and acoustic features for classification. 
Our results on two benchmark datasets reveal that acoustic features are helpful in improving the overall accuracy. 
Finally, a deeper analysis shows that acoustic features are valuable in three cases: when a dialog act has sufficient data, when lexical information is limited and when strong lexical cues are not present.
\end{abstract}
\section{Introduction}
\label{sec:intro}
\vsp

Every utterance in a conversation has a level of illocutionary force \cite{austin1962tw} whose meaning induces an effect over the course of the dialog. That meaning can be categorized into dialog acts (DAs) taking into account the relationship between the words being used and the force of the utterance \cite{book:kentBach2000}. 
A DA is the expression of the speaker's attitude or intention at every utterance in a conversation. Kent Bach \cite{book:kentBach2000} illustrates this by pointing out that a statement expresses a belief, a request expresses a desire, and an apology expresses a regret. In this manner, dialogs can be studied and modeled by analyzing their sequence of DAs.

Automatic DA tagging is an important preprocessing step for semantic extraction in natural language understanding and dialog systems. This task has been approached using two main information sources: lexical cues from dialog transcripts and acoustic cues from speech signals. 
For the former, traditional statistical algorithms have been employed, such as hidden Markov models (HMMs) \cite{Stolcke:2000}, conditional random fields (CRFs), \cite{Zimmermann:CRF2009} and support vector machines (SVMs) \cite{Henderson:SVM2010}. 
Recently, deep learning (DL) techniques, such as convolutional neural networks (CNNs) \cite{RCNN:KalchbrennerB13, lee:CNN_RNN_DA_2016}, recurrent neural networks (RNNs) \cite{lee:CNN_RNN_DA_2016,DBLP:journals/corr/JiHE16} and long short-term memory (LSTM) models \cite{AttentionCNN:ShenL16},
 have attained the state-of-the-art results in DA classification.

DAs can be ambiguous if only lexical information is considered. For example, a \textit{Declarative Question} like "\textit{this is your car(?)}" is hard to distinguish from a \textit{Statement} if the question mark is not present, and can easily be misclassified due to word order. In this case, acoustic information can help disambiguate. Moreover, in real applications that involve automatic speech recognition (ASR), a DA classifier can help deal with noisy transcriptions. Hence, some researchers  \cite{Stolcke:2000, DBLP:shiberg2000} have explored acoustic and prosodic cues from the speech signal as a potential knowledge source for DA classification.

Other works \cite{Arsikere2016, Ondas2015} have showed improvements exploring combinations of lexico-acoustic features. Inspired by these works, we present a neural hybrid model that takes both lexical and acoustic features as input in order to classify dialog utterances into DAs. Our model is a combination of two neural-based models: one, which processes lexical features of the utterances and their context (based on \cite{ortega-vu:2017:SIGDIAL}), and a second, which processes acoustic features. 
Our experiments show that acoustic features are helpful for improving overall accuracy and attaining state-of-the-art results on two benchmark datasets: the ICSI Meeting Recorder Dialog Act Corpus (MRDA) and the NXT-format Switchboard Corpus (SwDA). We also include an analysis of the acoustic features contribution for DA classification in three circumstances: when a DA has sufficient data, when strong lexical cues are missing and for single-word utterances.
\vsp

\section{Model}
\label{sec:model}
\begin{figure}[ht]
\begin{minipage}[h]{\linewidth}
  \centering
  \centerline{\includegraphics[width=8.5cm]{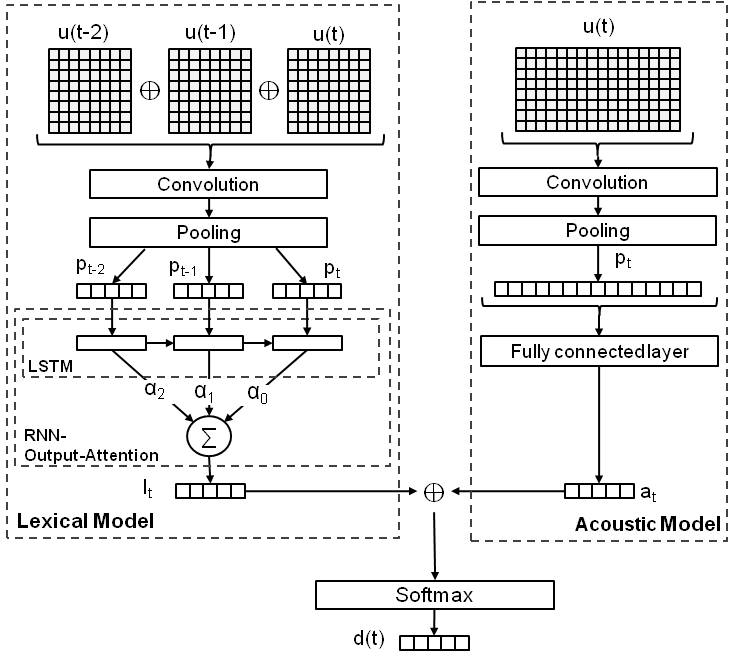}}
\end{minipage}
\caption{Architecture of the lexico-acoustic model. $\oplus$ represents a concatenation.}
\label{fig:LAM}%
\end{figure}

\vsp
The architecture of the lexico-acoustic model (LAM) proposed in this paper is depicted in Figure \ref{fig:LAM}. 
It contains two main parts: on the left side is the lexical model (LM) and on the right side the acoustic model (AM). Models are detailed in sections \ref{sec:LM}-\ref{sec:LAM}.

\vsp
\subsection{Lexical model}\label{sec:LM}

The LM, based on \cite{ortega-vu:2017:SIGDIAL}, takes the concatenation of grid-like representations of the current utterance and its $n$ previous utterances in the dialog as input to be processed by a CNN, generating a vector representation for each of those utterances. 

The CNN performs a discrete convolution using a set of different filters on an input matrix, where each column of the matrix is the word embedding of the corresponding word.
We use 2D filters $f$ (with width $|f|$) spanning over all embedding dimensions $d$ as described by the following equation:
\begin{equation}
 (w \ast f)(x,y) = \sum_{i=1}^{d}\sum_{j = -|f|/2}^{|f|/2}w(i,j) \cdot f(x-i,y-j)
\end{equation}

After convolution, an utterance-wise max pooling operation is applied. 
Then, the feature maps are concatenated, resulting in one vector per utterance.
These are represented in Figure \ref{fig:LAM} as $p_{t-2}, p_{t-1}$ and $p_t$.

The vector representations of the utterances are then processed by a context learning method, the RNN-Output-Attention (ROA) proposed in \cite{ortega-vu:2017:SIGDIAL}, in order to model the relation between each utterance and its context. 
ROA consists of an RNN with LSTM units followed by a weighted sum of the RNN's hidden states using an attention mechanism  \cite{Attn_ML:Bahdanau2014}. 


For each of the hidden state vector $h(t-i)$ at time step $t-i$ in a dialog, where $t$ is the current time step. The attention weights $\alpha_{i} $ are computed as follows

\begin{equation}
\alpha_i = \frac{exp(f(h(t-i)))}{\sum_{j} {exp(f(h(t-j))}}
\end{equation}
where $f$ is the scoring function. 
In our work, $f$ is the linear function of the input $h(t-i)$
\begin{equation}
f(h(t-i))= W^T h(t-i)
\end{equation}
where $W$ is a trainable parameter.
The output $l_t$ is the weighted sum of the hidden states sequence.
\begin{equation}
l_t= \sum_{i} \alpha_{i} h(t-i)
\end{equation}

Finally, the context representation $l_t$ is fed into a softmax layer that outputs a probability distribution over the DA set, given the current dialog utterance.
\vsp
\subsection{Acoustic model}\label{sec:AM}
We propose a CNN-based model to process acoustic features, because the speech signal of the utterances encodes important information for DA classification that is not contained in the transcripts.
The acoustic features from the speech signal are not taken at word or utterance level, but at frame level, i.e. the speech signal is divided into frames of 25 ms with a shift of 10 ms, and 13 Mel-frequency-cepstral coefficients (MFCC) per frame are extracted using the openSMILE toolkit \cite{tools:openSMILE}. 
MFCC features are stacked sequentially in order to obtain a grid-like input representation of the acoustic signal. 

The input is processed by a one-layer CNN using filters that span over the 13 MFCC features and 5 frames a time, with a max pooling layer in order to obtain a fixed-length vector representation.
This is fed into a softmax layer for DA classification, as explained in Section \ref{sec:LM}. 

\vsp
\subsection{Lexico-acoustic model}\label{sec:LAM}

The core of this work is the LAM (depicted in Figure \ref{fig:LAM}), a bi-CNN that employs lexical and acoustic cues simultaneously as input. 
The LAM combines a LM and a AM by concatenating the vector representations ($l_t$ and $a_t$) obtained from the context processing method in the LM and the pooling layer in the AM. 
Both vectors represent the current utterance, and can therefore be joined at this level and passed to the softmax function to output a final probability distribution over the DA set.

\vsp
\section{Experiments}
\label{sec:experiments}
\vsp
\subsection{Data}
We test our model on two DA datasets:
1) \textbf{MRDA}: ICSI Meeting Recorder Dialog Act Corpus \cite{corpus:ICSI_annotated}, a dialog corpus of \textit{multiparty meetings}. The 5-tag-set used in this work was introduced by \cite{ICSI:tagsetAng}.
2) \textbf{SwDA}: NXT-format Switchboard Corpus \cite{NXT:2010}, a dialog corpus of \textit{2-speaker conversations}.

NXT-format Switchboard Corpus was preferred over the original Switchboard Dialog Act Corpus \cite{Corpus:Switchboard,jurafsky97switchboard} because the former provides utterance transcripts and DA annotations as well as the time stamps at word level that were useful to extract acoustic features. Nonetheless, this corpus only provides DA annotation for roughly 50\% of the original dataset.

Train, validation and test splits on MRDA were taken as defined in \cite{lee:CNN_RNN_DA_2016}. However, on SwDA the splits were built by taking the annotated conversations from NXT-format Switchboard Corpus that appear in the split lists published in \cite{lee:CNN_RNN_DA_2016}. The new train, validation and test splits are roughly the half of the conversations on each original split. Summary statistics are shown in Table \ref{tab:datasets}. In both datasets, the classes are highly unbalanced; the majority class is $59.1 $\% on MRDA and $34.7$~\% on SwDA.

\begin{table}[ht]
\centering
\begin{tabular}{|l|c|c|c|c|c|}
\hline \textbf{Dataset}& \textbf{C}& \textbf{$\mid$V$\mid$}& \textbf{Train}& \textbf{Validation}&\textbf{Test}\\ \hline
MRDA	&5	&12k	& 78k	&16k	&15k\\
SwDA	&42	&16k	&98k	&8.5k	&2.5k\\
\hline
\end{tabular}
\caption{\label{tab:datasets} Data statistics: \textbf{C} is the number of classes, \textbf{$\mid$V$\mid$} is the vocabulary size and Train/Validation/Test are \#utterances.}
\end{table}

\vsp
\subsection{Hyperparameters and Training}
The hyperparameters of the three models for both datasets are summarized in Table \ref{tab:hyperparams}. The LM's hyperparameters were taken from \cite{ortega-vu:2017:SIGDIAL}, while the AM's hyperparameters were obtained by varying one hyperparameter at a time while keeping the others fixed. 
Training was done for 25 epochs with averaged stochastic gradient descent \cite{ASGD:Polyak1992} over mini-batches. 
The learning rate was initialized at 0.11 and reduced 10\% every 2000 parameter updates. Word2vec pretrained embeddings \cite{WordEmbeddings:word2vec} were employed and tuned during training. The context length $n$ was taken from the original the LM, i.e. $n=3$ for MRDA and $n=2$ for SwDA.

\begin{table}[ht]
\centering
\resizebox{\textwidth}{!}{\begin{minipage}{1.05\textwidth}
\begin{tabular}{|l|c|c|}
\hline 
\textbf{Hyperparameter}		&\textbf{LM} &\textbf{AM}\\ 
\hline
    Filter width 				&3, 4, 5 & 5\\
    Feature maps per filter		&100 &100 \\
    Dropout rate				&0.5 &0.5 \\
    Activation function 		&ReLU&ReLU  \\
    Pooling size		 		&utterance-wise&(18,1)\\
    Word embeddings				&word2vec\cite{WordEmbeddings:word2vec} (dim. 300)&--- \\
    MFCC features				&---&13 \\
    Mini-batch size				&\multicolumn{2}{c|}{50 (MRDA) -- 150 (SwDA)}  \\
\hline
\end{tabular}
\end{minipage}}
\caption{\label{tab:hyperparams} Hyperparameters. }
\end{table}

\vsp
\vsp
\subsection{Results}
\label{sec:Results}

Table \ref{tab:results} shows the results obtained from the three models on both datasets. As expected, the LM is superior to the AM, i.e. the lexical features yield more valuable information than the acoustic features for our task. 
On both datasets, the LM's accuracy is significantly higher than the AM's accuracy. 
However, for both datasets, the combined model yields improvements over both constituent models.
It indicates that both cue sources complement each other. 
\begin{table}[ht]
  \centering
  \begin{tabular}{|l|c|c|}
      \hline
      \textbf{Model} & \textbf{MRDA} & \textbf{SwDA} \\
      \hline
      Lexical		  &84.1 & 73.6\\
      Acoustic   &67.8  & 50.9  \\
      Lexico-acoustic &\textbf{84.7}& \textbf{75.1} \\
      \hline
  \end{tabular}%
  \caption{Accuracy (\%) of the three models on both datasets.}
  \label{tab:results}
\end{table}

\vsp
\section{Analysis}
\label{sec:analysis}
\vsp
This section's goal is to analyze the impact of joining both models, and to report and discuss which DAs benefit and which are impacted negatively, by applying a LAM versus a LM. Moreover, we also investigate the effect of the acoustic features when the question mark (\textit{?}) is removed from transcripts  and when utterances are very short.  

On MRDA, as reported in the previous section, the LAM yielded an improvement of 0.6\% over the LM. However, the improvement is not uniform over the five classes. While the prediction of the DAs \textit{Statement}, \textit{Disruption} and \textit{Backchannel} obtains a benefit from the acoustic features, \textit{Filter} is impacted negatively and \textit{Question} stays the same. Nonetheless, in general terms, the LAM benefits the overall DA classification, specially for those DAs with a higher presence in the training set, and the degradation caused by the model does not hurt its overall performance. 

On SwDA, the LAM also outperformed the LM by 1.5\%. Five DAs benefited by adding acoustic features: \textit{Statement}, \textit{Backchannel}, \textit{Opinion}, \textit{Abandon} and \textit{Agree}, \textit{Wh\_question} and \textit{Acknowledge} were negatively affected in a minimal extent, and the remaining 35 DAs were not impacted. These results are again highly correlated to the DA distribution in the corpus -- the 5 most frequent DAs obtained an improvement that is reflected in the overall accuracy. Therefore, we argue that the LAM helps when a large number of examples per DA is available. 
One possible reason is that we have enough training data for these particular DAs to properly train the AM part of the LAM.

\paragraph*{Effect of removing the question mark}\par
Contrary to our initial hypothesis that acoustic features would improve the accuracy of classifying \textit{Question}, no improvement was noted. 
Therefore, we analyzed how the LM and the LAM performed on this particular DA more deeply. 
The question mark \textit{?} in the manual transcripts plays a fundamental role for the DA \textit{Question} in the LM; 97.7\% of the utterances with question marks which are labeled as \textit{Question} are correctly predicted (see Table \ref{tab:question}) by the LM. For that reason, the acoustic features are not able to provide any useful information.

Consequently, we retrained and tested the LM and the LAM using transcripts from which the question mark was removed. 
This change also makes the transcripts more similar to transcripts from an ASR, where punctuation is not available or is not highly accurate. 
As expected, the overall accuracy dropped, from 84.1\% to 80.8\% in the LM and from 84.7\% to 81.9\% in the LAM. Although both models were affected by this modification, the LAM performed 1.1\% better than the LM, versus the improvement of 0.6\% with the original transcripts. Acoustic features slightly dampen the negative effect on the accuracy of removing the question mark.

Table \ref{tab:question} shows the accuracy of the LM and the LAM exclusively on utterances whose DA is \textit{Question} and which have a question mark in the manual transcript. 
The second column corresponds to the models which were trained and tested on the original transcripts and the third column to the models which were trained on transcripts with question marks removed. 
As mentioned above, the LM has a high accuracy at correctly predicting \textit{Question} if the utterance has the question mark. 
Moreover, when the acoustic features are added, the accuracy decreases by 1.6\%. 
Nonetheless, if question marks are not present in the data, the LM's accuracy drops to 46.6\%. 
This shows that this character is the most important cue at lexical level. The LAM's accuracy drops to 50.2\%, but this time it is superior to the LM by 3.6\%. 
This indicates that acoustic information is an important source of cues for tasks that use DA classification over data that lacks these important lexical cues, such as spoken language understanding.

\begin{table}[ht]
\centering
\begin{tabular}{|l|c|c|}
\hline
\textbf{Model} & \textbf{With \textit{'?'}} & \textbf{\textit{'?'} removed}  \\
\hline
Lexical 		&\textbf{97.7}	&46.6		\\
Lexico-acoustic &96.1  	& \textbf{50.2} 	\\
\hline
\end{tabular}%
\caption{Accuracy (\%) of \textit{Question} utterances on MRDA with question mark and when the question mark is removed.}
\label{tab:question}
\end{table}

\vspace{-0.5em}
\paragraph*{Single-word utterances}\par 
There exist utterances like \textit{Right} or \textit{Yeah} that are very common across several DAs. One of their characteristics is that they are very short and consequently they do not yield much information for classification. \cite{ortega-vu:2017:SIGDIAL,lee:CNN_RNN_DA_2016, Liu:DA} have successfully explored the use of context as a way to differentiate these type of utterances. In line with these works, both the LM and the LAM (in its lexical component) encode the context. 

We have shown in Section \ref{sec:Results} that the LAM outperforms the LM on both datasets, however, we explored particularly the effect of using acoustic features on the utterances \textit{Right} and \textit{Yeah} that are frequently tagged as \textit{Statement} and \textit{Backchannel} on MRDA. For our analysis purposes, we extracted the predictions of the utterances that exclusively contained one word that is either \textit{Right} or \textit{Yeah}, from which we can artificially define four subclasses: \textit{Statement-Right}, \textit{Backchannel-Right}, \textit{Statement-Yeah} and \textit{Backchannel-Yeah}. 

Table \ref{tab:class_right} shows the precision, recall and F\textsubscript{1} score of the LM and the LAM for the utterances \textit{Right}. 
On the one hand, for the DA \textit{Statement} the LAM achieves a higher F\textsubscript{1} score than the LM, while on the other hand, the F\textsubscript{1} score for \textit{Backchannel} decreases slightly. 
This means that using acoustic features improves the classification of utterances \textit{Right} as \textit{Statement} without affecting those utterances tagged as \textit{Backchannel}. 
A similar phenomenon is observed with  utterances \textit{Yeah}, however, in this case, the LAM improves the F\textsubscript{1} score for both DAs \textit{Statement} and \textit{Backchannel} (see Table \ref{tab:class_yeah}).
\begin{table}[ht]
\centering
\begin{tabular}{|l|l|c|c|c|}
\hline
\textbf{DA-Right} &\textbf{Model} & \textbf{P} & \textbf{R} & \textbf{F\textsubscript{1}}\\
\hline
\multicolumn{1}{|l|}{\multirow{2}{*}{Statement}}	&Lexical	&0.62	&0.35	&0.45	\\
\multicolumn{1}{|l|}{} 	&Lexico-acoustic	&0.60  	&0.45 	&\textbf{0.52}	\\
\hline
\multicolumn{1}{|l|}{\multirow{2}{*}{Backchannel}}   	&Lexical	&0.56  	&0.85 	&\textbf{0.67}	\\
\multicolumn{1}{|l|}{}   	&Lexico-acoustic	&0.56  	&0.77 	&0.65 	\\
\hline
\end{tabular}%
\caption{Precision, recall and F\textsubscript{1} score for the utterances \textit{Right}}
\label{tab:class_right}
\end{table}

\vsp
\begin{table}[ht]
\centering
\begin{tabular}{|l|l|c|c|c|}
\hline
\textbf{DA-Yeah} &\textbf{Model} & \textbf{P} & \textbf{R} & \textbf{F\textsubscript{1}}\\
\hline
\multicolumn{1}{|l|}{\multirow{2}{*}{Statement}}	&Lexical	&0.65	&0.36	&0.46	\\
\multicolumn{1}{|l|}{} 	&Lexico-acoustic	&0.67  	&0.50 	&\textbf{0.57}	\\
\hline
\multicolumn{1}{|l|}{\multirow{2}{*}{Backchannel}}   	&Lexical&0.60  	&0.89 	&0.72 	\\
\multicolumn{1}{|l|}{}   	&Lexico-acoustic	&0.64  	&0.87 	&\textbf{0.74} 	\\
\hline
\end{tabular}%
\caption{Precision, recall and F\textsubscript{1} score for the utterances \textit{Yeah}}
\label{tab:class_yeah}
\end{table}


\section{Comparison with other works}
\label{sec:comparison}
\vsp
We present a comparison between different works and our model in Table \ref{tab:comparison}. On MRDA, as we used the setup proposed by \cite{lee:CNN_RNN_DA_2016}, our results can only be compared accurately to \cite{ortega-vu:2017:SIGDIAL} and \cite{lee:CNN_RNN_DA_2016}, and the LAM outperforms both works. 
On SwDA, as we used the data available in the NXT format, and, to the best of our knowledge, no other model has been trained and tested on this subset of SwDA, our results cannot be strictly compared with other works.

\begin{table}[ht!]
\small
\centering
\begin{tabular}{|l|c|c|}
    \hline 
    \textbf{Model}	& \textbf{MRDA}	& \textbf{SwDA}    \\ 
    \hline
	LAM	(Our model)	&\textbf{84.7 }&\textbf{75.1}\\
    NCRL 			&84.3 &73.8 \\
    CNN-FF 			&84.6 &73.1 \\
    HBM  	 		&81.3 &---\\
    CNN+DAP         &---  &79.9\\
    HCNN 			&---  &73.9 \\
    HMM 			&---  &71.0 \\
    Majority class	&59.1 &34.7 \\
\hline
\end{tabular}
\label{tab:comparison}
\caption{Comparison of accuracy (\%).
\textit{NCRL}: Neural context representation learning proposed in \cite{ortega-vu:2017:SIGDIAL},
\textit{CNN-FF}: proposed in \cite{lee:CNN_RNN_DA_2016},
\textit{HBM}: hidden backoff model \cite{HiddenBackoff:Ji2006}. 
\textit{CNN+DAP}:proposed by \cite{Liu:DA}.  
\textit{HCNN}: hierarchical CNN \cite{RCNN:KalchbrennerB13}.
\textit{HMM} \cite{Stolcke:2000}. 
\textit{Majority class} is the most frequent class.
}
\end{table}
\vspace{-1.3em}
\section{Conclusion}
\label{ssec:conclusion}
\vsp
We proposed an approach to incorporate lexical and acoustic features in a neural model for DA classification. Our experiments on two benchmark datasets reveal that adding acoustic information to the model improves the overall accuracy attaining state-of-the-art results. 
A deeper analysis showed that acoustic features specially help when the data for a particular DA is large enough, when lexical information is limited, as in single-word utterances, and when strong lexical cues are not present.  
\vsp
\section{Acknowledgements}
\label{ssec:Ak}
\vsp
This work was funded by the National Council of Science and Technology of Mexico (CONACyT), the German Academic Exchange Service (DAAD) and the German Science Foundation (DFG), Sonderforschungsbereich 732, Project A8, at the University of Stuttgart.

\bibliographystyle{IEEEbib}
\bibliography{ICASSP}

\begin{thebibliography}{10}

\bibitem{austin1962tw}
J.L. Austin,
\newblock {\em {How to Do Things with Words}},
\newblock 1962.

\bibitem{book:kentBach2000}
Kent Bach,
\newblock {\em Concise Routledge Encyclopedia of Philosophy}, chapter Speech
  Acts,
\newblock 2000.

\bibitem{Stolcke:2000}
Andreas Stolcke, Noah Coccaro, Rebecca Bates, Paul Taylor, Carol Van
  Ess-Dykema, Klaus Ries, Elizabeth Shriberg, Daniel Jurafsky, Rachel Martin,
  and Marie Meteer,
\newblock ``Dialogue act modeling for automatic tagging and recognition of
  conversational speech,''
\newblock {\em Comput. Linguist.}, 2000.

\bibitem{Zimmermann:CRF2009}
Matthias Zimmermann,
\newblock ``Joint segmentation and classification of dialog acts using
  conditional random fields.,''
\newblock in {\em INTERSPEECH}, 2009.

\bibitem{Henderson:SVM2010}
M.~Henderson, M.~Gašić, B.~Thomson, P.~Tsiakoulis, K.~Yu, and S.~Young,
\newblock ``Discriminative spoken language understanding using word confusion
  networks,''
\newblock in {\em IEEE SLT}, 2012.

\bibitem{RCNN:KalchbrennerB13}
Nal Kalchbrenner and Phil Blunsom,
\newblock ``Recurrent convolutional neural networks for discourse
  compositionality,''
\newblock {\em CoRR}, 2013.

\bibitem{lee:CNN_RNN_DA_2016}
Ji~Young Lee and Franck Dernoncourt,
\newblock ``Sequential short-text classification with recurrent and
  convolutional neural networks,''
\newblock {\em CoRR}, 2016.

\bibitem{DBLP:journals/corr/JiHE16}
Yangfeng Ji, Gholamreza Haffari, and Jacob Eisenstein,
\newblock ``A latent variable recurrent neural network for discourse relation
  language models,''
\newblock {\em CoRR}, 2016.

\bibitem{AttentionCNN:ShenL16}
Sheng{-}syun Shen and Hung{-}yi Lee,
\newblock ``Neural attention models for sequence classification: Analysis and
  application to key term extraction and dialogue act detection,''
\newblock {\em CoRR}, 2016.

\bibitem{DBLP:shiberg2000}
Elizabeth Shriberg, Rebecca~A. Bates, Andreas Stolcke, Paul Taylor, Daniel
  Jurafsky, Klaus Ries, Noah Coccaro, Rachel Martin, Marie Meteer, and
  Carol~Van Ess{-}Dykema,
\newblock ``Can prosody aid the automatic classification of dialog acts in
  conversational speech?,''
\newblock {\em CoRR}, 2000.

\bibitem{Arsikere2016}
H.~Arsikere, A.~Sen, A.~P. Prathosh, and V.~Tyagi,
\newblock ``Novel acoustic features for automatic dialog-act tagging,''
\newblock in {\em 2016 IEEE International Conference on Acoustics, Speech and
  Signal Processing (ICASSP)}, 2016.

\bibitem{Ondas2015}
Stanislav Ondáš and Jozef Juhár,
\newblock ``Distance-based dialog acts labeling,''
\newblock in {\em Cognitive Infocommunications (CogInfoCom), 2015 6th IEEE
  International Conference on}, 2015.

\bibitem{ortega-vu:2017:SIGDIAL}
Daniel Ortega and Ngoc~Thang Vu,
\newblock ``Neural-based context representation learning for dialog act
  classification,''
\newblock in {\em SIGdial}, 2017.

\bibitem{Attn_ML:Bahdanau2014}
Dzmitry Bahdanau, Kyunghyun Cho, and Yoshua Bengio,
\newblock ``Neural machine translation by jointly learning to align and
  translate,''
\newblock {\em CoRR}, 2014.

\bibitem{tools:openSMILE}
Florian Eyben, Felix Weninger, Florian Gross, and Bj\"{o}rn Schuller,
\newblock ``Recent developments in opensmile, the munich open-source multimedia
  feature extractor,''
\newblock in {\em ACM International Conference on Multimedia}, 2013.

\bibitem{corpus:ICSI_annotated}
Elizabeth Shriberg, Raj Dhillon, Sonali Bhagat, Jeremy Ang, and Hannah Carvey,
\newblock ``The icsi meeting recorder dialog act (mrda) corpus,''
\newblock in {\em SIGdial}, 2004.

\bibitem{ICSI:tagsetAng}
Jeremy Ang, Yang Liu, and Elizabeth Shriberg,
\newblock ``Automatic dialog act segmentation and classification in multiparty
  meetings,''
\newblock in {\em ICASSP}, 2005.

\bibitem{NXT:2010}
Sasha Calhoun, Jean Carletta, Jason~M. Brenier, Neil Mayo, Dan Jurafsky, Mark
  Steedman, and David Beaver,
\newblock ``The nxt-format switchboard corpus: A rich resource for
  investigating the syntax, semantics, pragmatics and prosody of dialogue,''
\newblock {\em Lang. Resour. Eval.}, 2010.

\bibitem{Corpus:Switchboard}
John~J. Godfrey, Edward~C. Holliman, and Jane McDaniel,
\newblock ``Switchboard: Telephone speech corpus for research and
  development,''
\newblock in {\em ICASSP}, 1992.

\bibitem{jurafsky97switchboard}
D.~Jurafsky, E.~Shriberg, and D.~Biasca,
\newblock ``{Switchboard SWBD-DAMSL shallow-discourse-function annotation
  coders manual},''
\newblock Tech. {R}ep., 1997.

\bibitem{ASGD:Polyak1992}
B.~T. Polyak and A.~B. Juditsky,
\newblock ``Acceleration of stochastic approximation by averaging,''
\newblock {\em SIAM J. Control Optim.}, 1992.

\bibitem{WordEmbeddings:word2vec}
Tomas Mikolov, Ilya Sutskever, Kai Chen, Greg Corrado, and Jeffrey Dean,
\newblock ``Distributed representations of words and phrases and their
  compositionality,''
\newblock {\em CoRR}, 2013.

\bibitem{Liu:DA}
Yang Liu, Kun Han, Zhao Tan, and Yun Lei,
\newblock ``Using context information for dialog act classification in dnn
  framework,''
\newblock in {\em EMNLP}, 2017.

\bibitem{HiddenBackoff:Ji2006}
Gang Ji and Jeff Bilmes,
\newblock ``Backoff model training using partially observed data: Application
  to dialog act tagging,''
\newblock in {\em HLT-NAACL}, 2006.

\end{thebibliography}

\end{document}